\documentclass{article}
\usepackage{ijcai21}

\usepackage{times}
\usepackage{soul}
\usepackage{url}
\usepackage[hidelinks]{hyperref}
\usepackage[utf8]{inputenc}
\usepackage[small]{caption}
\usepackage{graphicx}
\usepackage{amsmath}
\usepackage{amsthm}
\usepackage{booktabs}
\usepackage{algorithm}
\usepackage{algorithmic}
\usepackage{mathtools}
\usepackage{booktabs}
\usepackage{multirow}
\usepackage[inline]{enumitem}
\usepackage{url}

\newtheorem{theorem}{Theorem}
\newtheorem{lemma}{Lemma}
\newtheorem{corollary}{Corollary}
\newtheorem{proposition}{Proposition}

\theoremstyle{definition}
\newtheorem{Question}{Q}
\newtheorem{Answer}{A}
\def\!{\mskip-\thinmuskip}
\newcommand*{\ldblbrace}{\left \{\mskip-5mu \left  \{\,}
\newcommand*{\rdblbrace}{\, \right \}\mskip-5mu \right\}}

\title{UniGNN: a Unified Framework for Graph and Hypergraph Neural Networks 
}

\author{
Jing Huang$^1$
\and
Jie Yang$^{1}$\thanks{Corresponding author: Jie Yang. This work is partly supported by NSFC, China (No: 61876107, U1803261,  61806125).}
\affiliations
$^1$Institute of Image Processing and Pattern Recognition, Shanghai Jiao Tong University\\
\emails
\{ziruochenxia, jieyang\}@sjtu.edu.cn
}

\begin{document}

\maketitle

\begin{abstract}
    Hypergraph, an expressive structure with flexibility to model the higher-order correlations among entities, has recently attracted increasing attention from various research domains.  
    Despite the success of Graph Neural Networks (GNNs) for graph representation learning, how to adapt the powerful GNN-variants directly into hypergraphs remains a challenging problem. 
    In this paper, we propose UniGNN, a unified framework for interpreting the message passing process in graph and hypergraph neural networks, which can generalize general GNN models into hypergraphs. 
    In this framework, meticulously-designed architectures aiming to deepen GNNs can also  be incorporated into hypergraphs with the least effort.
    Extensive experiments have been conducted to demonstrate the effectiveness of UniGNN on multiple real-world datasets, which outperform the state-of-the-art approaches with a large margin. 
    Especially for the DBLP dataset, we increase the accuracy from 77.4\% to 88.8\% 
    in the semi-supervised hypernode classification task.
    We further prove that the proposed message-passing based UniGNN models are at most as powerful as the 1-dimensional Generalized Weisfeiler-Leman (1-GWL) algorithm in terms of distinguishing non-isomorphic hypergraphs. 
    Our code is available at \url{https://github.com/OneForward/UniGNN}.
\end{abstract}

\section{Introduction}




Hypergraphs are natural extensions of graphs by allowing an edge to join any number of vertices, which can represent the higher-order relationships involving multiple entities. 
Recently, hypergraphs have drawn the attention from a wide range of fields, like computer vision \cite{gao2020hypergraph,ijcai2020-109:Hyper3DPose}, recommendation system \cite{dhcn_aaai21} and natural sciences \cite{Gu_2020:HyperQuantum}, and been incorporated with various domain-specific tasks. 

In a parallel note, Graph Representation Learning has raised a surge of interests from researchers. Numerous powerful Graph Neural Networks (GNNs) have been presented, achieving the state-of-the-art in specific graph-based tasks, such as node classification \cite{Chen2020:GCNII}, link prediction \cite{Zhang2018:SEAL} and graph classification \cite{Li2020:GXN}. Most GNNs are message-passing based models, like GCN \cite{Kipf2017:GCN}, GAT \cite{Velickovic2017:GAT}, GIN \cite{Xu2018:GIN} and GraphSAGE \cite{Hamilton2017:GraphSAGE}, 
which iteratively update node embeddings by aggregating neighboring nodes' information.
The expressive power of GNNs is well-known to be upper bounded by 1-Weisfeiler-Leman (1-WL) test \cite{Xu2018:GIN} and many provably more powerful GNNs mimicking higher-order-WL test have been presented \cite{Morris2019:kLWL,Morris2019-aaai:kGNN,Li2020a:DE-GNN}.

Furthermore, several works, like JKNet \cite{Xu2018:JKNet}, DropEdge \cite{Rong2019:DropEdge}, DGN \cite{Zhou2020a:DeepDiffGNorm} and GCNII \cite{Chen2020:GCNII}, 
have devoted substantial efforts to tackling the problem of \textit{over-smoothing}, an issue when node embeddings in GNNs tend to converge as layers are stacked up and the performance downgrades significantly.  



\begin{figure}[t]
    \centering
    \includegraphics[width=0.48\textwidth]{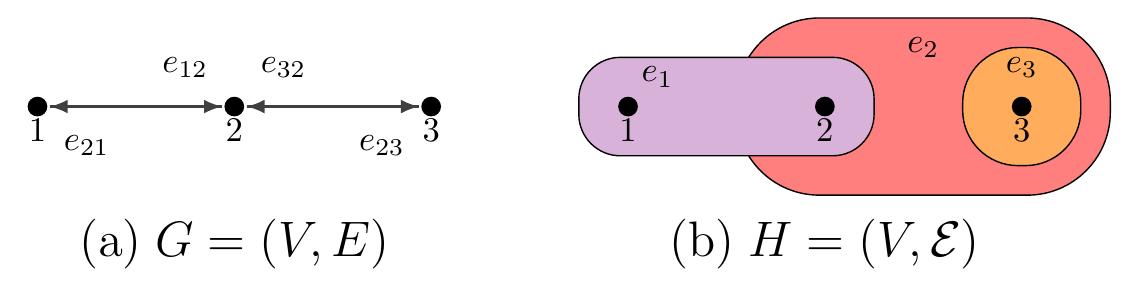}
    \caption{Toy examples of a graph and a hypergraph. (a) A graph $G$. (b) A hypergraph $H$. Note that $G$ can be reduced from $H$ using clique expansion.}
    \label{fig:1}
    \end{figure}

Despite the success of GNNs, how to learn powerful representative embeddings for hypergraphs remains a challenging problem. 
HGNN \cite{Feng2019:HGNN} is the first hypergraph neural network, which uses the clique expansion technique to approximate hypergraphs as graphs, and simplifies the problem above  to the graph embedding framework. This approach, as illustrated in Fig~\ref{fig:1}, cannot cover the substructures like hyperedges which recursively contain other hyperedges are discarded with clique expansion. 
HyperGCN \cite{Yadati2019:HyperGCN} enhances the generalized hypergraph Laplacian with additional weighted pairwise edges (a.k.a mediators). This approach still fails to reserve complete hypergraph information since Graph Laplacian can only describe pairwise connections between vertices in one training epoch.
Another work, HyperSAGE \cite{Arya2020:HyperSAGE}  learns to embed  hypergraphs directly by propagating messages in a two-stage procedure. Although HyperSAGE shows the capability to capture information from hypergraph structures with a giant leap in performance, it fails to adapt powerful classic GNN designs into hypergraphs.


In view of the fact that more and more meticulously-designed network architectures and learning strategies have appeared in graph learning, we are naturally motivated to ask the following intriguing question:

\textit{Can the network design and learning strategy for GNNs be applied to HyperGNNs directly?}

\paragraph{Contributions}
This paper proposes the UniGNN, a unified framework for graph and hypergraph neural networks, 
with contributions 
unfolded by the following questions:

\begin{Question} 
    Can we generalize the well-designed GNN architecture for hypergraphs with the least effort?
\end{Question}

\begin{Question} 
    Can we utilize the learning strategies for circumventing the over-smoothing in the graph learning and design deep neural networks that adapt to hypergraphs? 
\end{Question}

\begin{Question} 
    How powerful are hypergraph neural networks?
\end{Question}

By addressing the above questions, we highlight our contributions as follows:
\begin{Answer}
    We present the UniGNN and use it to generalize several classic GNNs, like GCN, GAT, GIN and GraphSAGE directly into hypergraphs, termed UniGCN, UniGAT, UniGIN and UniSAGE, respectively. UniGNNs consistently outperform the state-of-art approaches  in hypergraph learning tasks.
\end{Answer}

\begin{Answer}
    We propose the UniGCNII, the first deep hypergraph neural network  and verify its effectiveness in resolving the over-smoothing issue.
\end{Answer}

\begin{Answer}
    We prove that message-passing based UniGNNs are at most as powerful as 1-dimensional Generalized Weisfeiler-Leman (1-GWL)  algorithm in terms of distinguishing non-isomorphic hypergraphs.
\end{Answer}

\section{Preliminaries}

\subsection{Notations}

Let $G=(V, E)$ denote a directed or undirected graph consisting of a vertex set $V=\{1, \ldots, n\}$ and an edge set $E$ (pairs of vertices).
A \textit{self-looped graph} $\tilde{G}=(V, \tilde{E})$ is constructed from $G$ by adding a self-loop to each of its non-self-looped nodes. 
The \textit{neighbor-nodes} of vertex $i$ is denoted by $\mathcal{N}_i = \{  j | (i, j) \in E \} $.
We also denote vertex $i$'s neighbor-nodes with itself as $\tilde{\mathcal{N}_i} = \mathcal{N}_i  \bigcup \{ i \} $ .   
We use $x_i \in \mathbf{R}^d$ to represent a $d$-dimensional feature of vertex $i$.

A hypergraph $H=(V, \mathcal{E} )$ is defined as a generalized graph by allowing an edge to connect any number of vertices, where $V$ is a set of vertices and a \textit{hyperedge} $e \in \mathcal{E}$ is a non-empty subset of $V$.
The \textit{incident-edges} of vertex $i$ is denoted by $E_i = \{ e \in \mathcal{E} | i \in e \} $.
We say two hypergraphs $H_1=(V_1, \mathcal{E}_1)$ and $H_2=(V_2, \mathcal{E}_2)$ are \textit{isomorphic}, written $H_1 = H_2$,  if there exists a bijection $f: V_1 \to  V_2$ such that 
$\forall e = \{ v_1, \ldots, v_k \} \subseteq V_1: e \in \mathcal{E}_1 \Leftrightarrow  \{ f(v_1), \ldots, f(v_k) \} \in \mathcal{E}_2$.

\subsection{Graph Neural Networks}


\paragraph{General GNNs}
Graph Neural Networks (GNNs) learn the informative embedding of a graph by utilizing the feature matrix and the graph structure. 
A broad range of GNNs can be built up by the message passing layers, in which node embeddings are updated by aggregating the information of its neighbor embeddings. 
The message passing process in the $l$-th layer of a GNN is formulated as
\begin{equation}
    \label{eq:GNN-MP}
    \text{(GNN)}\;
    x^{l+1}_i = \phi^l \left( x^l_i , \{ x^l_j \}_{j \in \mathcal{N}_i} \right).
\end{equation}
Classic GNN models sharing this paradigm include GCN \cite{Kipf2017:GCN}, GAT \cite{Velickovic2017:GAT}, GIN \cite{Xu2018:GIN}, GraphSAGE \cite{Hamilton2017:GraphSAGE}, etc..


In the following sections, we omit the superscript $l$ for the sake of simplicity 
and use $\tilde{x}_i \in \mathbf{R}^{d'}$ to indicate the output of the message passing layer before activation or normalization. 



\subsection{HyperGraph Neural Networks}


\paragraph{Spectral-based HyperGNNs} 
HGNN  \cite{Feng2019:HGNN} and HyperConv \cite{Bai19:HyperConv}  utilize the normalized hypergraph Laplacian, which essentially converts the hypergraphs to conventional graphs by viewing each hyperedge as a complete graph.
HyperGCN \cite{Yadati2019:HyperGCN}  uses the generalized hypergraph Laplacian (changing between epochs) and injects information of mediators to represent hyperedges. 
Both methods depend on the hypergraph Laplacian, which however, emphasizes the pairwise relations between vertices.
Another work, MPNN-R \cite{Yadati2020:G-MPNN} regards hyperedges as new vertices with $\hat{V}=V \cup \mathcal{E} $ and represents the hypergraph by a $|\hat{V}| \times |\mathcal{E}|$ matrix. MPNN-R can effectively capture the recursive property of hyperedges, but fails to describe other high-order relationships, like complex and diverse intersections between hyperedges.

\paragraph{Spatial-based HyperGNNs}  A recent work, HyperSAGE \cite{Arya2020:HyperSAGE} pioneers to exploit the structure of hypergraphs by aggregating messages in a two-stage procedure, avoiding the information loss due to the reduction of hypergraphs to graphs. With  $\mathcal{N}(i, e)$ to denote vertex $i$'s \textit{intra-edge neighborhood} for hyperedge $e$, HyperSAGE aggregates information with the following rules:
\begin{equation}
    \label{eq:HyperSAGE}
    \begin{cases}
        h_{i,e} = \mathcal{M}_1 \left(  \{ x_j \}_{j \in  \mathcal{N}(i, e; \alpha) } \right) \\
        \tilde{x}_i = W \left( x_i + \mathcal{M}_2 ( \{  h_{i,e} \}_{e \in E_i} ) \right)
    \end{cases},
\end{equation}
where $\mathcal{N}(i, e; \alpha)$ is a sampled subset of $\alpha$ vertices from $\mathcal{N}(i, e)$, $W$ is the linear transform and  $\mathcal{M}_1$ and $\mathcal{M}_2$ are power mean functions.

HyperSAGE is the current state-of-the-art algorithm for hypergraph representation learning. However, there are some issues associated with HyperSAGE. 
Firstly, since the calculation for $h_{i, e}$ is distinct for different $(i, e)$ pairs, the original algorithm uses nested loops over hyperedges and vertices within hyperedges, which results in redundant computation and poor parallelism. 
Secondly, applying the power mean functions in both stages, neither of which is injective, fails to distinguish structures with the same distribution but different multiplicities of elements \cite{Xu2018:GIN}.
Lastly, the original work still fails to address the over-smoothing issue associated with deep hypergraph neural networks.


\section{UniGNN: a Unified Framework}

\begin{figure}[t]
    \centering
    \includegraphics[width=0.48\textwidth]{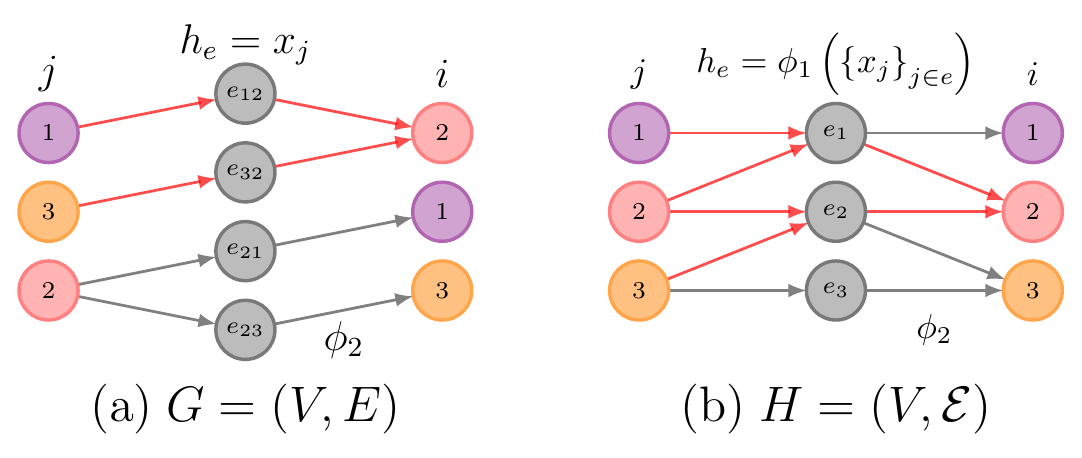}
    \caption{An illustration of how the UniGNN can be applied to Graphs (left) and Hypergraphs (right). (a) Two-stage message passing for graph $G$ in Figure \ref{fig:1}. (b) Two-stage message passing for hypergraph $H$ in Figure \ref{fig:1}. Note that edges showing how messages flow to vertex 2 are marked in red.} 
    \label{fig:2}
    \end{figure}

To resolve the issues associated with HyperSAGE, 
we propose the UniGNN, a unified framework to characterize the message-passing process in GNNs and HyperGNNs:
\begin{equation}
    \label{eq:UniGNN}
    \text{(UniGNN)}
    \begin{cases}
        h_e = \phi_1  \bigl( \{ x_j \}_{j \in e} \bigr)  \\
        \tilde{x}_i =  \phi_2 \Bigl( x_i,  \{ h_e \}_{e \in E_i}  \Bigr)
    \end{cases},
\end{equation}
where $\phi_1$ and $\phi_2$  are permutation-invariant functions for aggregating messages from vertices and hyperedges respectively. The update rule for UniGNN is illustrated in Fig~\ref{fig:2}.
The key insight is that if we rethink Eq.~\eqref{eq:GNN-MP} in GNNs as a two-stage aggregation process, then the designs for GNNs can be naturally generalized to hypergraphs in Eq.~\eqref{eq:UniGNN}.


In the first stage, for each hyperedge $e$, we use $\phi_1$ to aggregate features of all vertices within it. $\phi_1$ can be any permutation-invariant function satisfying
$\phi_1 \left( \left\{ x_j \right\} \right) = x_j$, such as the mean function $h_e = \frac{1}{|e|} \sum_{ j \in e }  x_j$ or the sum function $h_e = \sum_{ j \in e }  x_j$. It is obvious that if we let ${E}_i= \left\{ e=\{ j \} \; | \; \forall j \in {\mathcal{N}}_i \right\}$, then $h_e=x_j$ holds for any $j \in {\mathcal{N}}_i$. Therefore, UniGNN \eqref{eq:UniGNN} can be reduced to GNN \eqref{eq:GNN-MP}, which unifies both formulations into the same framework.

In the second stage, we update each vertex with its incident hyperedges using aggregating function $\phi_2$,  of which the design  can be inspired from existent GNNs directly. 
We will exhibit several effective examples  in the following section. 

\subsection{Generalize Powerful GNNs for Hypergraphs}
\label{subsection:shallowUniGNN}

\paragraph{UniGCN} 
Graph Convolutional Networks (GCN) \cite{Kipf2017:GCN} propagate the features using the weighted sum (where weights are specified by the node degrees), 
\begin{equation}
    \label{eq:GCN}
    \tilde{x}_i = \frac{1}{\sqrt{d_i}} \sum_{j \in \tilde{\mathcal{N}}_i} {\frac{1}{\sqrt{d_j}} W x_j},
\end{equation}
where $d_i = |\tilde{\mathcal{N}}_i|, \forall i \in V$. 

Based on our framework, we can generalize the above aggregation process to hypergraphs as 
\begin{equation}
    \label{eq:UniGCN}
    \tilde{x}_i = \frac{1}{\sqrt{d_i}} \sum_{e \in \tilde{E}_i} \frac{1}{\sqrt{d_e}} W h_e,
\end{equation}
where we define $d_e = \frac{1}{|e|} \sum_{i \in e} d_i$ as the \textit{average degree of a hyperedge} $e$. UniGCN endorses less weight to high-degree hyperedges in aggregation.
It is trivial that by letting $\tilde{E}_i= \left\{ e | e=\{ j \}, j \in \tilde{\mathcal{N}}_i \right\}$, then $d_e=d_j, h_e=x_j$, and thus UniGCN is reduced to GCN.

\paragraph{UniGAT} Graph Attention Networks (GAT) \cite{Velickovic2017:GAT} adopt the attention mechanism to assign importance score to each of center node's neighbors, leading to more effective aggregation. The attention mechanism is formulated as 
\begin{align}
    \alpha_{ij} &= \sigma ( a^T \left[ W x_i; W x_j \right] ), \\ 
    \tilde{\alpha}_{ij} &= \frac{ \exp{(\alpha_{ij})} } { \sum_{k \in \tilde{\mathcal{N}}_i} { \exp{(\alpha_{ik})}  } }, \\
    \tilde{x}_i &= \sum_{j \in \tilde{\mathcal{N}}_i} \tilde{\alpha}_{ij} W x_j, 
\end{align}
where $\sigma$ is the leaky ReLU function, $a \in \mathbf{R}^{2d'}$ is the learnable attentional parameter and $[\cdot; \cdot]$ means concatenation.

By rewriting the above equations, we can get UniGAT for hypergraphs  as follows,
\begin{align}
    \alpha_{ie} &= \sigma ( a^T \left[ W h_{\{i\}}; W h_e \right] ) ,\\ 
    \tilde{\alpha}_{ie} &= \frac{ \exp{(\alpha_{ie})} } { \sum_{e' \in \tilde{E}_i} { \exp{(\alpha_{ie'})}  } }, \\
    \tilde{x}_i &= \sum_{e \in \tilde{E}_i} \tilde{\alpha}_{ie} W h_e. 
\end{align}
In this way, UniGAT learns to reweight the center node's neighboring hyperedges.

UniGAT is essentially different from HyperGAT~\cite{Bai19:HyperConv} since HyperGAT requires the hyperedges to be preprocessed into the same homogeneous domain of vertices before training, which is inflexible and unreliable in practice.

Note that based on the formulation for UniGCN and UniGAT, hypergraphs should  be preprocessed with self-loops; that is, $ \forall i \in V \Rightarrow e=\{i\} \in \tilde{\mathcal{E}}$.

\paragraph{UniGIN} Graph Isomorphism Networks (GIN) \cite{Xu2018:GIN} is a simple yet effective model with the expressive power achieving the upper bound of message passing based GNNs. GIN updates node embeddings as 
\begin{equation}
    \tilde{x}_i = W \Big( (1 + \varepsilon ) x_i + \sum_{j \in \mathcal{N}_i} {x_j} \Big),
\end{equation}
where $\varepsilon$ is a learnable parameter or some fixed scalar.

Similar to the previous deduction, UniGIN is formulated as 
\begin{equation}
    \tilde{x}_i = W \Big( (1 + \varepsilon ) x_i + \sum_{e \in E_i} {h_e} \Big).
\end{equation}

\paragraph{UniSAGE}
GraphSAGE \cite{Hamilton2017:GraphSAGE} uses a general aggregating function, like mean aggregator, LSTM aggregator or max-pooling aggregator, which can be designed according to various tasks. We use a variant of GraphSAGE where the combining process is \textit{sum} instead of \textit{concatenation} following \cite{Arya2020:HyperSAGE}:
\begin{equation}
    \tilde{x}_i = W \left( x_i + {\text{AGGREGATE}}\left( \{ x_j \}_{j \in { \mathcal{N}}_i }  \right) \right).
\end{equation}

UniSAGE is naturally generalized as
\begin{equation}
    \tilde{x}_i = W \left( x_i + {\text{AGGREGATE}}\left( \{ h_e \}_{e \in E_i} \right) \right).
\end{equation}



\subsection{Towards Deep Hypergraph Neural Networks}

Current hypergraph representation learning methods, like HGNN, HyperGCN and HyperSAGE, use a shallow network with two layers and the performance reduces significantly when layers are stacked up, which is in concordance with GNNs. This phenomenon is called \textit{over-smoothing}.
Although many works have focused on tackling this problem for graphs, like JKNet \cite{Xu2018:JKNet}, DropEdge \cite{Rong2019:DropEdge}  and GCNII \cite{Chen2020:GCNII}, how to make hypergraphs deeper is still uncovered.


Since in our framework, learning strategies from graph learning domain can be incorporated into hypergraphs with the least effort, we solve this problem by presenting UniGCNII, a deep hypergraph neural network inspired from GCNII.

\paragraph{UniGCNII} GCNII \cite{Chen2020:GCNII} is a powerful deep graph convolutional network enhanced with \textit{Initial Residual Connection} and \textit{Identity Mapping} to vanquish the \textit{over-smoothing} problem. We generalize GCNII to hypergraphs, dubbed UniGCNII, with the aggregation process defined as
\begin{equation}
    \begin{cases}
        \hat{x}_i  = \frac{1}{\sqrt{d_i}} \sum_{e \in \tilde{E}_i} \frac{1}{\sqrt{d_e}} h_e \\ 
        \tilde{x}_i = \left((1 - \beta) I + \beta W\right) \left((1-\alpha)  \hat{x}_i + \alpha x^0_i\right) 
    \end{cases},
\end{equation}
where $\alpha$ and $\beta$ are hyperparameters, $I$ is identity matrix and $x^0_i$ is the initial feature of vertex $i$.


In each layer, UniGCNII employs the same two-stage aggregation as UniGCN to exploit the hypergraph structure, and then injects the jumping knowledge from the initial features and previous features. Experiments validate that UniGCNII  enjoys the advantage of circumventing the over-smoothing issue when models are getting deeper. 



\section{How Powerful are UniGNNs?}

Message-passing based GNNs are capable of distinguishing local-substructure (like $k$-height subtree rooted at a node) or global structure of graphs, with the expressive power upper bounded by 1-WL test.
In view of this, we are motivated to investigate the expressive power of UniGNNs for hypergraphs. We start by presenting a variant of the 1-dimensional Generalized Weisfeiler-Leman Algorithm (1-GWL) for hypergraph isomorphism test following the work of \cite{Boker2019:HyperColor}.

\subsection{Generalized Weisfeiler-Leman Algorithm}

1-GWL sets up by labeling the vertices of a hypergraph $H=(V, \mathcal{E})$ with $l^0_i = 0$ for any $i \in V$, and in the $t$-th iteration the labels are updated by
\begin{equation}
    \begin{cases}
    l^t_e =  \ldblbrace l^t_j \rdblbrace_{j \in e }, \forall e \in \mathcal{E} \\
    l^{t+1}_i = \ldblbrace (l^t_i, l^t_e) \rdblbrace_{e \in E_i}, \forall i \in V
    \end{cases},
\end{equation}
where  $l^t_e$ denotes the label of a hyperedge $e \in \mathcal{E}$, and $\ldblbrace \ldots \rdblbrace$ denotes a multiset.

1-GWL distinguish $H_1$ and $H_2$ as non-isomorphic if there exists a $t \ge 0$ such that
\begin{equation}
    \ldblbrace l^t_{H_1, i} \, | \, i \in V_1 \rdblbrace \neq \ldblbrace  l^t_{H_2, i} \, | \, i \in V_2 \rdblbrace,
\end{equation}
where the subscript $H_1$ and $H_2$ are added for discrimination.

\begin{proposition}[1-GWL] \label{prop:1-GWL}
    If 1-GWL test decides $H_1$ and $H_2$ are non-isomorphic, then $H_1 \neq H_2$.
\end{proposition}
We leave all the proofs in the supplemental files.

\subsection{Discriminative Power of UniGNNs}
We assign the same features to all vertices of a hypergraph $H$ so that the UniGNN only depends on the hypergraph structure to learn.
Let $\mathcal{A} : \mathcal{H} \mapsto \mathbf{R}^{g}$ be a UniGNN abiding by the aggregation rule~\eqref{eq:UniGNN}, the following proposition indicates that $\mathcal{A}$'s expressive power is upper bounded by 1-GWL test.

\begin{proposition}\label{prop:upper-bound}
    Given two non-isomorphic hypergraphs $H_1$ and $H_2$, if $\mathcal{A}$ can distinguish them by $\mathcal{A}(H_1) \neq \mathcal{A}(H_2)$, then 1-GWL test also decides $H_1 \neq H_2$.
\end{proposition}

The following theorem characterizes the conditions for UniGNNs to reach the expressive power of 1-GWL test.

\begin{theorem} \label{thm1:condition_global}
    Given two hypergraphs $H_1$ and $H_2$ such that 1-GWL test  decides as non-isomorphic,  a UniGNN  $\mathcal{A}$ is suffice to distinguish them by $\mathcal{A}(H_1) \neq \mathcal{A}(H_2)$ with the following conditions:
    \begin{enumerate}
        \item \textbf{Local Level.} Two-stage aggregating functions $\phi_1$ and $\phi_2$ are both injective.
        \item \textbf{Global Level.}  In addition to the local-level conditions, $\mathcal{A}$'s graph-level READOUT function is injective. 
    \end{enumerate}
\end{theorem}

We are also interested in UniGNNs' capability of distinguishing local substructures of hypergraphs. We define the \textit{local substructure} of a hypergraph $H$  as the $k$-height subtree of its \textit{incidence graph} $I(H)$, where  $I(H)$ is the bipartite graph  with vertices $V(I (H)) \coloneqq  V \cup \mathcal{E}$ and edges $E(I(H)) \coloneqq \{ ve \; | \; v \in e \; \text{for} \; e \in \mathcal{E} \}$. 


\begin{corollary} \label{cor:condition_local}
    Assume that 1-GWL test can distinguish two distinct local substructures from hypergraphs, the UniGNN  $\mathcal{A}$ can also distinguish them as long as the Local Level condition is satisfied.
\end{corollary}


\section{Experiments}

In this section, we evaluate the performance of the proposed methods in extensive experiments. 

\paragraph{Datasets}
We use the standard academic network datasets: DBLP \cite{Rossi15:DBLP}, Pubmed, Citeseer and Cora \cite{Sen_AI:Cora}   for all the experiments. The hypergraph is created with each vertex representing a document.
The co-authorship hypergraphs, constructed from DBLP and Cora, connect all documents co-authored by one author as one hyperedge.
The co-citation hypergraphs are built with PubMed, Citeseer and Cora, using one hyperedge to represent all documents cited by an author.
We use the same preprocessed hypergraphs as HyperGCN, which are publicly available in their official implementation\footnote{\url{https://github.com/malllabiisc/HyperGCN}}.

\begin{table*}[h]
    \centering
    \begin{tabular}{@{}lccccc@{}}
    \toprule
                    & \multicolumn{2}{c}{\textbf{Co-authorship Data}} & \multicolumn{3}{c}{\textbf{Co-citation Data}}                   \\ 
                    \cmidrule(lr){2-3} \cmidrule(lr){4-6}
    \textbf{Method} & DBLP                   & Cora                   & Pubmed              & Citeseer            & Cora                \\
                    \cmidrule(lr){1-1} \cmidrule(lr){2-2} \cmidrule(lr){3-3} \cmidrule(lr){4-4} \cmidrule(lr){5-5} \cmidrule(lr){6-6} 
    MLP+HLR         & 63.6 ± 4.7             & 59.8 ± 4.7             & 64.7 ± 3.1          & 56.1 ± 2.6          & 61.0 ± 4.1          \\
    HGNN            & 69.2 ± 5.1             & 63.2 ± 3.1             & 66.8 ± 3.7          & 56.7 ± 3.8          & \textbf{70.0 ± 2.9} \\
    FastHyperGCN    & 68.1 ± 9.6             & 61.1 ± 8.2             & 65.7 ± 11.1         & 56.2 ± 8.1          & 61.3 ± 10.3         \\
    HyperGCN        & 70.9 ± 8.3             & 63.9 ± 7.3             & 68.3 ± 9.5          & 57.3 ± 7.3          & 62.5 ± 9.7          \\
    HyperSAGE       & 77.4 ± 3.8             & 72.4 ± 1.6             & 72.9 ± 1.3          & 61.8 ± 2.3          & 69.3 ± 2.7          \\ \midrule
    UniGAT          & \textbf{88.7 ± 0.2}    & 75.0 ± 1.1             & \textbf{74.7 ± 1.2} & \textbf{63.8 ± 1.6} & 69.2 ± 2.9          \\ 
    UniGCN          & \textbf{88.8 ± 0.2}    & \textbf{75.3 ± 1.2}    & 74.4 ± 1.0          & 63.6 ± 1.3          & \textbf{70.1 ± 1.4} \\
    UniGIN          & 88.6 ± 0.3             & 74.8 ± 1.3             & 74.4 ± 1.1          & 63.3 ± 1.2          & 69.2 ± 1.5          \\
    UniSAGE         & 88.5 ± 0.2             & 75.1 ± 1.2             & 74.3 ± 1.0          & \textbf{63.8 ± 1.3} & \textbf{70.2 ± 1.5} \\ \bottomrule
    \end{tabular}
    \caption{Testing accuracy (\%) of UniGNNs and other hypergraph models on co-authorship and co-citation datasets for \textit{Semi-supervised Hypernode Classification}. The best or competitive results are highlighted for each dataset.}
    \label{tab:Semi}
\end{table*}

\begin{table*}[h]
    \centering
    \begin{tabular}{@{}lcccccccc@{}}
    \toprule
              & \multicolumn{2}{c}{DBLP}      & \multicolumn{2}{c}{Pubmed}    & \multicolumn{2}{c}{Citeseer}  & \multicolumn{2}{c}{Cora(cocitation)} \\ 
    \textbf{Method}    & seen          & unseen        & seen          & unseen        & seen          & unseen        & seen     & unseen           \\ 
              \cmidrule(lr){1-1} \cmidrule(lr){2-3} \cmidrule(lr){4-5} \cmidrule(lr){6-7} \cmidrule(lr){8-9}
    MLP+HLR   & 64.5        & 58.7          & 66.8          & 62.4          & 60.1          & 58.2          & 65.7          & 64.2          \\
    HyperSAGE & 78.1        & 73.2          & 81.0          & 80.4          & 69.3          & 67.9          & 71.3          & 66.8          \\ \midrule
    UniGAT  & 88.4          & 82.7          & 83.5          & 83.4          & 70.9          & \textbf{71.3} & 72.4          & 70.1          \\ 
    UniGCN  & 88.5          & 82.6          & 83.7          & 83.3          & 71.2          & 70.6          & \textbf{74.3} & 71.5          \\
    UniGCN* & 88.4          & 82.8          & \textbf{85.7} & \textbf{85.1} & 68.2          & 70.6          & 74.1          & \textbf{71.8} \\
    UniGIN  & \textbf{89.6} & \textbf{83.4} & 83.8          & 83.3          & \textbf{71.5} & 70.8          & 73.7          & 71.3          \\
    UniSAGE & 89.3          & 83.0          & 83.6          & 83.1          & 71.1          & 70.8          & \textbf{74.2} & 71.5          \\ \bottomrule
    \end{tabular}
    \caption{Testing accuracies(\%) of multiple UniGNN variants and other hypergraph learning methods for \textit{Inductive Learning on Evolving Hypergraphs}. The best or competitive results are highlighted for each dataset.}
    \label{tab:Evolving}
\end{table*}

\begin{table}[h]
    
    \small
    \setlength\tabcolsep{4.7pt}
    \begin{tabular}{@{}clllllll@{}}
    \toprule
    \multirow{2}{*}{Dataset}    & \multicolumn{1}{c}{\multirow{2}{*}{Models}} & \multicolumn{6}{c}{Layers}                                                  \\ 
                                & \multicolumn{1}{c}{}                        & \multicolumn{1}{c}{2} & \multicolumn{1}{c}{4} & \multicolumn{1}{c}{8} & \multicolumn{1}{c}{16} & \multicolumn{1}{c}{32} & \multicolumn{1}{c}{64}    \\ \midrule
    \multirow{5}{*}{DBLP}                                                      & UniGAT            & \textbf{89.1} & 66.4          & 21.2          & 16.2 & 16.2          & OOM           \\ 
                                                                               & UniGCN            & \textbf{89.2} & 79.2          & 18.6          & 16.2 & 16.2          & 16.2          \\
                                                                               & UniGIN            & \textbf{89.6} & 88.3          & 47.9          & 26.6 & 23.1          & 16.3          \\
                                                                               & UniSAGE           & \textbf{89.4} & 88.2          & 46.7          & 31.0 & 20.6          & 16.2          \\
                                                                               & \textbf{UniGCNII} & 88.4          & 87.6          & 88.4          & 89.3 & 89.3          & \textbf{89.4} \\ \midrule
    \multirow{5}{*}{\begin{tabular}[c]{@{}c@{}}Cora\\ Coauthor\end{tabular}}   & UniGAT            & \textbf{76.0} & 64.0          & 31.7          & 29.4 & 29.1          & 30.4          \\
                                                                               & UniGCN            & \textbf{76.2} & 68.7          & 38.2          & 28.7 & 29.2          & 29.4          \\
                                                                               & UniGIN            & \textbf{75.8} & 68.3          & 39.0          & 28.3 & 28.5          & 30.2          \\
                                                                               & UniSAGE           & \textbf{75.9} & 68.7          & 37.4          & 28.9 & 28.3          & 28.6          \\
                                                                               & \textbf{UniGCNII} & 75.1          & 74.2          & 75.1          & 76.1 & \textbf{76.6} & 76.5          \\ \midrule
    \multirow{5}{*}{Pubmed}                                                    & UniGAT            & \textbf{75.2} & 68.8          & 61.9          & 55.8 & 41.1          & 39.7          \\
                                                                               & UniGCN            & \textbf{74.9} & 73.7          & 61.2          & 49.5 & 41.7          & 39.8          \\
                                                                               & UniGIN            & \textbf{74.8} & 73.6          & 60.6          & 49.7 & 41.6          & 40.3          \\
                                                                               & UniSAGE           & \textbf{74.8} & 73.3          & 61.6          & 50.2 & 41.5          & 39.7          \\
                                                                               & \textbf{UniGCNII} & 75.6          & \textbf{75.8} & \textbf{75.8} & 75.4 & 75.4          & 75.4          \\ \midrule
    \multirow{5}{*}{Citeseer}                                                  & UniGAT            & \textbf{65.4} & 51.9          & 33.9          & 27.2 & 21.3          & 19.9          \\
                                                                               & UniGCN            & \textbf{64.5} & 58.7          & 35.5          & 23.3 & 21.0          & 20.2          \\
                                                                               & UniGIN            & \textbf{64.6} & 59.3          & 36.9          & 27.0 & 21.9          & 20.0          \\
                                                                               & UniSAGE           & \textbf{65.0} & 59.0          & 36.6          & 26.8 & 21.4          & 20.6          \\
                                                                               & \textbf{UniGCNII} & 64.1          & 63.3          & 63.9          & 65.8 & 66.4          & \textbf{66.5} \\ \midrule
    \multirow{5}{*}{\begin{tabular}[c]{@{}c@{}}Cora\\ Cocitation\end{tabular}} & UniGAT            & \textbf{70.9} & 55.4          & 30.7          & 27.9 & 23.6          & 27.3          \\
                                                                               & UniGCN            & \textbf{71.2} & 62.9          & 31.5          & 25.3 & 26.7          & 27.2          \\
                                                                               & UniGIN            & \textbf{70.9} & 62.7          & 35.1          & 26.9 & 28.0          & 27.2          \\
                                                                               & UniSAGE           & \textbf{71.4} & 63.3          & 33.6          & 26.6 & 26.1          & 27.2          \\
                                                                               & \textbf{UniGCNII} & 70.0          & 70.4          & 72.3          & 73.4 & \textbf{73.6} & 73.3          \\ \bottomrule 
    \end{tabular}
    \caption{Testing accuracies(\%) of multiple UniGNN variants with different depths in \textit{Semi-supervised Hypernode Classification} task. The result of the best performed model for each dataset is bolded. Note that additional validation data are used in this experiment.}
    \label{tab:DeepSemi}
    \end{table}

\begin{table}[h]
    
    \setlength\tabcolsep{5.5pt}
    \begin{tabular}{@{}lcccc@{}}
    \toprule
             & \multicolumn{2}{c}{UniGAT}       & \multicolumn{2}{c}{UniGCN}       \\ \cmidrule(lr){2-3}  \cmidrule(lr){4-5}
             & w/o        & w/                  & w/o        & w/                  \\ \cmidrule(lr){2-2}  \cmidrule(lr){3-3} \cmidrule(lr){4-4}  \cmidrule(lr){5-5}
    DBLP     & 88.1 ± 0.1 & \textbf{88.7 ± 0.2} & 88.1 ± 0.1 & \textbf{88.8 ± 0.2} \\
    Cora~1    & 67.4 ± 1.5 & \textbf{75.0 ± 1.1} & 67.3 ± 2.0 & \textbf{75.3 ± 1.2} \\
    Pubmed   & 30.1 ± 0.8 & \textbf{74.7 ± 1.2} & 30.2 ± 0.9 & \textbf{74.4 ± 1.0} \\
    Citeseer & 39.8 ± 1.2 & \textbf{63.8 ± 1.6} & 40.2 ± 1.3 & \textbf{63.6 ± 1.3} \\
    Cora~2    & 43.8 ± 3.9 & \textbf{69.2 ± 2.9} & 44.1 ± 3.6 & \textbf{70.1 ± 1.4} \\ \bottomrule
    \end{tabular}
    \caption{Testing accuracies(\%) of UniGCN and UniGAT in \textit{Semi-supervised Hypernode Classification} task, when input hypergraphs are with or without self-loops. Cora~1 is for Cora coauthorship and Cora~2 is for Cora cocitation. }
    \label{tab:Selfloop}
\end{table}

\subsection{Semi-supervised Hypernode Classification}

\paragraph{Setting Up and Baselines.}
The semi-supervised hypernode classification task aims to predict labels for the test nodes, given the hypergraph structure, all nodes' features and very limited training labels. The label rate of each dataset can be found in the supplemental materials.

We employ four two-layer UniGNN variants: UniGCN, UniGAT, UniGIN and UniSAGE.
For all models, mean function is used as the first-stage aggregation.
UniSAGE uses the SUM function for the second-stage aggregation.
Note that as described in Section~\ref{subsection:shallowUniGNN}, hypergraphs are preprocessed with self-loops for UniGCN and UniGAT.

We compare UniGNN models against the following baselines: 
(a) Multi-Layer Perceptron with explicit Hypergraph Laplacian Regularization (\textbf{MLP+HLR}),  (b) HyperGraph Neural Network (\textbf{HGNN} \cite{Feng2019:HGNN}), (c) HyperGraph Convolutional Network (\textbf{HyperGCN} \cite{Yadati2019:HyperGCN}) and (d) \textbf{HyperSAGE} \cite{Arya2020:HyperSAGE}.

Closely following the previous works, for each model on each dataset, we repeat experiments over 10 data splits with 8 different random seeds, amounting to 80 experiments.  We use the Adam optimizer with a learning rate of 0.01 and the weight decay of 0.0005. We fix the training epochs as 200 and 
report the performance of the model of the last epoch. The same training/testing split as \cite{Yadati2019:HyperGCN} is used. 
We 
 run all experiments on a single NVIDIA 1080Ti(11GB).

\paragraph{Comparison to SOTAs.} Table \ref{tab:Semi} summarizes the mean classification accuracy with the standard deviation on on the test split of UniGNN variants after 80 runs.
 We reuse the metrics that are already reported in \cite{Arya2020:HyperSAGE} for MLP+HLR, HGNN, HyperGCN and the best metrics reported for HyperSAGE. 

Results in Table \ref{tab:Semi} demonstrate that  UniGNNs are consistently better than the baselines with a considerable lift, achieving a new state-of-the-art. 
Especially for the DBLP dataset, we significantly improve the accuracy from 77.4\% to 88.8\% with negligible variance. 
On all datasets, our results are generally more stable than the baselines, as indicated by lower standard deviation.
Whereas for the Cora cocitation dataset, we report only slight improvements against SOTA, we argue that this is due to the fact that the Cora cocitation hypergraph contains the least mean hyperedge size $|M|=3.0 \pm 1.1$, for which the information loss from clique expansion in HGNN might be negligible.

Overall, with the powerful aggregation designs inspired from GNNs, our UniGNN models can effectively capture the intrinsic structure information from hypergraphs and perform stably better prediction with less deviation. 

\paragraph{Effect of Self-loops for UniGCN and UniGAT} 
We further study the effect of self-loops for UniGCN and UniGAT. 
Table \ref{tab:Selfloop} reports the mean accuracies for UniGCN and UniGAT when input hypergraphs are with or without self-loops. We observe that when hypergraphs are un-self-looped, the performances drop significantly for most datasets, which support the correctness of the formulations in Section~\ref{subsection:shallowUniGNN}.

\subsection{Inductive Learning on Evolving Hypergraphs}
\paragraph{Setting Up}
The task for inductive learning on evolving hypergraph takes the \textit{historical hypergraph} as input and predicts the unseen nodes' labels.

We closely follow \cite{Arya2020:HyperSAGE} and use the corrupted hypergraph which randomly removes 40\% vertices as unseen data during training. 20\%  vertices are used  for training and the rest 40\%   for the seen part of testing vertices. 
The other experimental settings are similar to those in the transductive semi-supervised learning task.
We employ an additional UniGCN variant, denoted as UniGCN*, which applies the linear transform after aggregation and normalization. 
We compare our models against MLP+HLR and HyperSAGE and use the best results reported from \cite{Arya2020:HyperSAGE}. 

\paragraph{Comparison to SOTAs.}
Table \ref{tab:Evolving} reports the mean classification accuracy on seen part and unseen part of the testing data. We observe that our UniGNN models consistently show better scores across the benchmark datasets. Similar to the semi-supervised setting, our models notably show significant improvements in dataset DBLP, where the prediction accuracy increases from 78.1\% to 89.6\% in the seen data and from 73.2\% to 83.4\% in the unseen data.

Results from table \ref{tab:Evolving} confirm that UniGNNs can capture the global structure information and perform well for predicting unseen nodes in the inductive learning task.

\subsection{Performance of Deep-layered UniGNNs}
\paragraph{Setting Up.}
To verify the effectiveness of UniGCNII, we study how the performance changes for  vanilla UniGNN models with various depths. In this experiment, we use the same setting as described in the semi-supervised hypernode classification task, except that additional 20\% of the original testing split is used as the validation split.

For UniGCNII, we perform all experiments in 1000 epochs and early stopping with a patience of 150 epochs. We use the Adam Optimizer with a learning rate of 0.01. We set the L2 regularizer factor to 0.01 for the  convolutional layers, 0.0005 for the dense layer, which is the same as described in GCNII \cite{Chen2020:GCNII}. Please refer to the supplemental materials for more details.


\paragraph{Comparison with Other Deep-layered UniGNNs.}
Table \ref{tab:DeepSemi} summarizes the results, in which the best performed model for each dataset is bolded. 
We see that UniGCNII enjoys the benefit of deep network structures and shows generally better results as layers increase.
We highlight that UniGCNII outperforms the best shallow models in dataset Cora, Pubmed and Citeseer, and obtains competitive results in dataset DBLP. 
On the contrary, the performance of vanilla models  drop significantly as depths increase.

Overall, the results suggest that our proposed framework is capable of incorporating meticulously-designed deep GNN models for deep hypergraph learning.

\section{Conclusion}

We propose the UniGNN, a unified framework for graph and hypergraph neural networks. Under this framework, we naturally generalize several classic GNNs to HyperGNNs,
which consistently show stably better performances than recent state-of-the-art methods.
We firstly solve the over-smoothing problem of deep hypergraph neural networks by presenting the UniGCNII.
Our models learn expressive representation of hypergraphs, which can be beneficial for a broad range of downstream tasks.
Future works include designing provably more powerful UniGNNs with high-order GWL test.
Another interesting direction for future work is to design hypergraph subtree kernel for hypergraph classification.



\newpage

\bibliographystyle{named}
\bibliography{ijcai21}

\appendix

\section{Proofs}

\subsection{Proof for Proposition~\ref{prop:1-GWL}}

\begin{proof}
    We use $l^t_{H}$ to denote the labels for $H$ in iteration $t$ and use $E_{H, i}$ to denote the incident hyperedges of vertex $i$ in $H$.

    We prove this proposition by contrapositive.
    Suppose $H_1=H_2$, then there exist a bijection $f: V_1 \to V_2$ and a bijection $g: \mathcal{E}_1 \to \mathcal{E}_2$ such that $\forall e = \{ v_1, \ldots, v_k \} \subseteq V_1: e \in \mathcal{E}_1 \Leftrightarrow g(e) = \{ f(v_1), \ldots, f(v_k) \} \in \mathcal{E}_2$.
    
    Since $H_1=H_2$ implies $|V_1|=|V_2|$, in iteration 0, we have 
    \[
        l^0_{H_1}= \ldblbrace l^0_{H_1, i}=0 \rdblbrace_{i \in V_1 } = \ldblbrace  l^0_{H_2, i}=0 \rdblbrace_{i \in V_2} =l^0_{H_2}.
    \]

    Assume that in iteration $t$, $l^t_{H_1}=l^t_{H_2}$ holds. Then in iteration $t+1$, we have 
    $$
    l^t_{H_1, e} =  \ldblbrace l^t_{H_1, j} \rdblbrace_{j \in e } = \ldblbrace l^t_{H_2, f(j)} \rdblbrace_{f(j) \in g(e) } =  l^t_{H_2, g(e)},
    $$
    \begin{align*}
        l^{t+1}_{H_1, i} &= \ldblbrace (l^t_{H_1, i}, l^t_{H_1, e}) \rdblbrace_{e \in E_{H_1, i}} \\ 
        &= \ldblbrace (l^t_{H_2, f(i)}, l^t_{H_2, g(e)}) \rdblbrace_{g(e) \in E_{H_2, f(i)}} \\
        &= l^{t+1}_{H_2, f(i)}.
    \end{align*}
    Note that $f$ is bijective; thus, we have $l^{t+1}_{H_1}=l^{t+1}_{H_2}$.  
    
    By induction rules, this proves that for any iteration $t$, $l^t_{H_1}=l^t_{H_2}$ holds and thereby the proposition.
\end{proof}

\subsection{Proof for Proposition~\ref{prop:upper-bound}}

We use $h^t_{H, i}$ to denote the feature of vertex $i$ in iteration $t$ and use $h^t_{H}$ to denote the multiset of all vertices' features in hypergraph $H$.

\begin{lemma}
    Let $[0:T]$ denote the set $\{ 0, \ldots, T \}$.
    If we have
    \begin{equation}
        \begin{cases}
            l^t_{H_1} = l^t_{H_2} & \forall t \in [0:T] \\ 
            h^t_{H_1} = h^t_{H_2} & \forall t \in [0:T-1]
        \end{cases},
        \label{eq:lemma}
    \end{equation}
    then $\forall t \in [0:T], \forall i_1 \in V_1, \forall i_2 \in V_2 $, we have 
    $$
    l^t_{H_1, i_1} = l^t_{H_2, i_2} \Rightarrow   h^t_{H_1, i_1} = h^t_{H_2, i_2}.
    $$
\end{lemma}

\begin{proof}[Proof of Lemma]
    In iteration $T=0$, we do not need the second equation in \eqref{eq:lemma}.  The lemma holds since 1-GWL test and UniGNN $\mathcal{A}$ start with the same vertex labels/features.

    Assume that $T > 0$ and $ \forall t \in [0:T-1]$, $l^t_{H_1, i_1} = l^t_{H_2, i_2} \Rightarrow   h^t_{H_1, i_1} = h^t_{H_2, i_2}$ holds. If in iteration $T$, we also have $l^T_{H_1, i_1} = l^T_{H_2, i_2}$, that is, 
    \begin{align*}
        &  \ldblbrace \left( l^{T-1}_{H_1, i_1}, \ldblbrace l^{T-1}_{H_1, j_1}  \rdblbrace _ {j_1 \in e_1}  \right) _ {e_1 \in E_{H_1, i_1}}  \rdblbrace = \\ 
        &  \ldblbrace \left( l^{T-1}_{H_2, i_2}, \ldblbrace l^{T-1}_{H_2, j_2}  \rdblbrace _ {j_2 \in e_2}  \right) _ {e_2 \in E_{H_2, i_2}}  \rdblbrace.
    \end{align*}

    Based on our assumption, we must have 
    \begin{align*}
        &  \ldblbrace \left( h^{T-1}_{H_1, i_1}, \ldblbrace h^{T-1}_{H_1, j_1}  \rdblbrace _ {j_1 \in e_1}  \right) _ {e_1 \in E_{H_1, i_1}}  \rdblbrace = \\ 
        &  \ldblbrace \left( h^{T-1}_{H_2, i_2}, \ldblbrace h^{T-1}_{H_2, j_2}  \rdblbrace _ {j_2 \in e_2}  \right) _ {e_2 \in E_{H_2, i_2}}  \rdblbrace.
    \end{align*}
    
    Note that UniGNN $\mathcal{A}$ witnesses the same neighborhood for vertex $i_1 \in V_1$  and vertex $i_2 \in V_2$; thus, two-stage aggregating functions generate the same output for them, i.e. $h^T_{H_1, i_1} = h^T_{H_2, i_2}$. Therefore, the lemma holds.
\end{proof}

\begin{proof}[Proof of Proposition~\ref{prop:upper-bound}]

    
    Suppose that there exists a minimal $T \ge 0$ such that after $T$ iterations, the UniGNN has  $\mathcal{A}(H_1) \neq \mathcal{A}(H_2)$, i.e. $h^T_{H_1} \neq h^T_{H_2}$, but  the 1-GWL test can not decide $H_1$ and $H_2$ as non-isomorphic, i.e. $l^T_{H_1} = l^T_{H_2}$.

    If $T=0$, then $l^T_{H_1} = l^T_{H_2}$ implies $h^T_{H_1} = h^T_{H_2}$ since 1-GWL test and UniGNN $\mathcal{A}$ start with the same vertex labels/features. This is contradictive to our assumption.

    If $T > 0$, we have $l^t_{H_1} = l^t_{H_2}$ and  $h^t_{H_1} = h^t_{H_2}$ for $t=0, \ldots, T-1$.  
    By assumption $l^T_{H_1} = l^T_{H_2}$, then the conditions for the lemma hold. Thus, we have 
    $$
        l^T_{H_1, i_1} = l^T_{H_2, i_2} \Rightarrow   h^T_{H_1, i_1} = h^T_{H_2, i_2}
    $$
    which indicates there exists a valid mapping $\mathcal{M}$ such that $ \mathcal{M}(l^t_{H, i}) \mapsto h^t_{H, i}$. Since $l^T_{H_1} = l^T_{H_2}$, then 
\begin{align*}
    &\ldblbrace l^T_{H_1, i_1}  \rdblbrace _{i_1 \in V_1}
    = \ldblbrace l^T_{H_2, i_2}  \rdblbrace _{i_2 \in V_2} \\ 
    &\Rightarrow \ldblbrace \mathcal{M}(l^T_{H_1, i_1} ) \rdblbrace _{i_1 \in V_1}
    = \ldblbrace \mathcal{M}(l^T_{H_2, i_2} ) \rdblbrace  _{i_2 \in V_2}
\end{align*}
Therefore, we have
\begin{align*}
    h^T_{H_1} &= \ldblbrace  h^T_{H_1, i_1} \rdblbrace _{i_1 \in V_1}
    = \ldblbrace \mathcal{M}(l^T_{H_1, i_1} ) \rdblbrace _{i_1 \in V_1} \\
    & = \ldblbrace \mathcal{M}(l^T_{H_2, i_2} ) \rdblbrace  _{i_2 \in V_2}
     = \ldblbrace h^T_{H_2, i_2} \rdblbrace  _{i_2 \in V_2}
     = h^T_{H_2}
\end{align*}
This results in the contradiction against our assumption and thereby we prove the original proposition.

\end{proof}

\subsection{Proof for Theorem~\ref{thm1:condition_global}}

\begin{proof}
    We start by proving that there exists an injective mapping 
    $\mathcal{M}^t$ such that $ \mathcal{M}^t (l^t_{H,i}) \mapsto h^t_{H,i}$.

    If $t=0$, it's trivial to prove that $\mathcal{M}^0$ exists since $l^0_{H,i}$ and $h^0_{H,i}$ are both initialized with the same labels/features.

    Assume that $t \ge 0$ and the injective mapping $\mathcal{M}^t$ exists, then for iteration $t+1$,
    \begin{align*}
        h^{t+1}_{H, i} = \phi_2 \left( \ldblbrace  \left(  h^t_{H,i},  \phi_1 \left( \ldblbrace h^t_{H,j}  \rdblbrace _{ j \in e }
          \right)  \right) \rdblbrace _{e \in E_i}  \right)
    \end{align*}
    Since the composition of injective functions $\phi_1, \phi_2$ is still injective, we can rewrite it as 
    \begin{align*}
        h^{t+1}_{H, i} &= \varphi \left( \ldblbrace  \left(  h^t_{H,i},   \ldblbrace h^t_{H,j}  \rdblbrace _{ j \in e }
           \right) \rdblbrace _{e \in E_i}  \right) \\ 
        &= \varphi \left( \ldblbrace  \left(  \mathcal{M}^t (l^t_{H,i}),   \ldblbrace \mathcal{M}^t  (l^t_{H,j})  \rdblbrace _{ j \in e }
           \right) \rdblbrace _{e \in E_i}  \right) \\ 
        &= \mathcal{M}^{t+1} \left( \ldblbrace  \left(  l^t_{H,i},   \ldblbrace l^t_{H,j}  \rdblbrace _{ j \in e }
           \right) \rdblbrace _{e \in E_i}  \right) \\ 
        &= \mathcal{M}^{t+1} \left( l^{t+1}_{H, i} \right), 
    \end{align*}
    where injective function $\varphi$ is induced from $\phi_1, \phi_2$ and injective function $\mathcal{M}^{t+1}$ is induced by $\varphi, \mathcal{M}^{t}$. 
    
    Therefore, by induction rule, there always exists an injective mapping $\mathcal{M}^t$ such that $ \mathcal{M}^t (l^t_{H,i}) \mapsto h^t_{H,i}$.

    Thus, if in iteration $T$, $l^T_{H_1} \neq l^T_{H_2}$, then 
    \begin{align*}
        \ldblbrace  l^T_{H_1, i} \rdblbrace _{i \in V_1} \neq
        \ldblbrace  l^T_{H_2, i} \rdblbrace _{i \in V_2}.
    \end{align*}
     Therefore, by injectivity, we have 
    \begin{align*}
        h^T_{H_1} &= \ldblbrace  h^T_{H_1, i}  \rdblbrace _{i \in V_1} = \ldblbrace \mathcal{M}^T ( l^T_{H_1, i} )  \rdblbrace _{i \in V_1} \\ 
        & \neq \ldblbrace \mathcal{M}^T ( l^T_{H_2, i} )  \rdblbrace _{i \in V_2} = \ldblbrace  h^T_{H_2, i}  \rdblbrace _{i \in V_2} = h^T_{H_2}.
    \end{align*}

    Since the global READOUT function is also injective, this implies that if 1-GWL test decides $H_1$ and $H_2$ as non-isomorphic, UniGNN $\mathcal{A}$ can also distinguish them by $\mathcal{A}(H_1) \neq \mathcal{A}(H_2)$.
\end{proof}

\subsection{Proof for Corollary~\ref{cor:condition_local}}

\begin{proof}
    We can remap the local substructure from incident graph $I(H)$ to a subhypergraph $H'$ and repeat the proof on the subhypergraph as described above. Since we don't need a global READOUT, the Local Level conditions are sufficient.
\end{proof}

\section{Implementation Details}

\subsection{Hyper-parameters Details}
Table~\ref{tab:HyperParameters} summarizes the model hyper-parameters and optimizers for all experiments.
$L_2$ represents the weight decay for shallow-UniGNNs.
$L_{2d}$  and $L_{2c}$ denote the weight decay for dense layer and convolutional layer of UniGCNII respectively.
For DBLP dataset, we remove the normalization layer for each model.


\begin{table}[h]
    \begin{tabular}{@{}ll@{}}
    \toprule
    Method  & Hyper-parameters \\ \midrule
    \multirow{2}{*}{UniGAT}   & hidden: 8, heads: 8,  dropout: 0.6,  \\    &lr: 0.01, $L_2$: 5e-4 , attentional dropout: 0.6           \\ \midrule
    \multirow{2}{*}{UniGCN}   & hidden: 64, dropout: 0.6, \\  &lr: 0.01, $L_2$: 5e-4       \\ \midrule
    \multirow{2}{*}{UniGIN}   & hidden: 64, dropout: 0.6, \\      &lr: 0.01, $L_2$: 5e-4          \\ \midrule
    \multirow{2}{*}{UniSAGE}  & hidden: 64,  dropout: 0.6, \\ & lr: 0.01, $L_2$: 5e-4, AGGREGATOR:SUM             \\  \midrule
    \multirow{2}{*}{UniGCNII}   & hidden: 64, dropout: 0.2, \\  & lr: 0.01, $L_{2c}$: 0.01,   $L_{2d}$: 5e-4      \\ \bottomrule
    \end{tabular}
    \caption{The model hyper-parameters for all experiments.}
    \label{tab:HyperParameters}
\end{table}

\subsection{Experimental Details}

\paragraph{Semi-supervised Hypernode Classification}
We train each model in each dataset to 200 epochs and report  the performance of the model of the last epoch.
The experiment of each model for each dataset is conducted 80 times with 10 train/test splits and 8 different random seeds.

\paragraph{Inductive Learning on Evolving Hypergraphs}
We train each model in each dataset with 200 epochs and report  the performance of the model of the last epoch.
For PubMed dataset, we set the input dropout as 0 instead of 0.6 and run 300 epochs. The experiment of each model for each dataset is conducted 80 times with 10 train/test splits and 8 different random seeds.

\paragraph{Performance of Deep-layerd UniGNNs}
In this experiment, additional 20\% of the original testing split is used as the validation split and the rest 80\% is used as the new testing split.
For the shallow UniGNNs, we train each model in each dataset in 200 epochs  and report the performance of the model with the highest validation score.
For UniGCNII, we perform all experiments in 1000 epochs and early stopping with a patience of 150 epochs.
The experiment of each model for each dataset is conducted 80 times with 10 train/val/test splits and 8 different random seeds.

\section{Details of Datasets}

\begin{table}[h]
    \begin{tabular}{@{}lrrrrrr@{}}
    \toprule
    Dataset  & \multicolumn{1}{c}{$|V|$} & \multicolumn{1}{c}{$|\mathcal{E}|$} & \multicolumn{1}{c}{$|M|$} 
             & \multicolumn{1}{c}{$d$} & \multicolumn{1}{c}{$C$} & \multicolumn{1}{c}{$\eta$} \\ \midrule
    DBLP     & 43413   & 22535    & 4.7 ± 6.1    & 1425      & 6       & 4.0\%     \\
    PubMed   & 19717   & 7963     & 4.3 ± 5.7    & 500       & 3       & 0.8\%     \\
    Citeseer & 3312    & 1079     & 3.2 ± 2.0    & 3703      & 6       & 4.2\%     \\
    Cora~1    & 2708    & 1072     & 4.2 ± 4.1    & 1433      & 7       & 5.2\%     \\
    Cora~2    & 2708    & 1579     & 3.0 ± 1.1    & 1433      & 7       & 5.2\%     \\ \bottomrule
    \end{tabular}
    \caption{Dataset Statistics. $|M|=\frac{1}{|\mathcal{E}|} \sum_{e \in \mathcal{E}} |e|$ denotes the average size of hyperedges.
     $d$, $C$ and $\eta$ denotes the dimension of input features, the number of classes and label rate, respectively. 
    Cora~1 is for Cora coauthorship and Cora~2 is for Cora cocitation. }
    \label{tab:dataset}
\end{table}

Statistics of the datasets are summarized in Table~\ref{tab:dataset}.

\end{document}